\documentclass[journal]{IEEEtran}

\ifCLASSINFOpdf
\else
\fi
\usepackage[table]{xcolor}
\usepackage{cite}
\usepackage{amsmath,amssymb,amsfonts}
\usepackage{subfigure}
\usepackage{graphicx}
\usepackage{algorithmic}
\usepackage{textcomp}
\usepackage[ruled,linesnumbered ]{algorithm2e}
\usepackage{xcolor}
\usepackage{mathtools}
\DeclarePairedDelimiter\abs{\lvert}{\rvert}
\usepackage{hyperref}
\usepackage{mathtools}
\usepackage{multirow}
\usepackage{tabularx}
\usepackage{arydshln}
\usepackage{dblfloatfix}

\makeatletter
\def\hlinewd#1{%
\noalign{\ifnum0=`}\fi\hrule \@height #1 %
\futurelet\reserved@a\@xhline}
\makeatother

\definecolor{ForestGreen}{RGB}{0, 155, 0}

\usepackage{bm}
\usepackage{tikz}

\usepackage{amssymb}

\let\emptyset\varnothing

\begin{document}

\newcommand{\Arrow}[1]{%
\parbox{#1}{\tikz{\draw[<-](0,0)--(#1,0);}}
}

\title{Continuous-Curvature Target Tree Algorithm for Path Planning in Complex Parking Environments}

\author{Minsoo Kim$^{1}$,
        Joonwoo Ahn$^{1}$,
        and Jaeheung Park$^{1, 2}$
\thanks{$^{1}$The authors are with DYROS Lab, Graduate School of Convergence Science and Technology, Seoul National University, Seoul, Republic of Korea. 
        {\tt\small (msk930512, joonwooahn, park73)@snu.ac.kr}}
\thanks{$^{2}$Jaeheung Park is also with Advanced Institutes of Convergence Technology(AICT), Suwon, Republic of Korea. He is the corresponding author of this paper}
\thanks{This work has been submitted to the IEEE for possible publication. Copyright may be transferred without notice, after which this version may no longer be accessible.}
}




\maketitle


\begin{abstract}
Rapidly-exploring random tree (RRT) has been applied for autonomous parking due to quickly solving high-dimensional motion planning and easily reflecting constraints.
However, planning time increases by the low probability of extending toward narrow parking spots without collisions.
To reduce the planning time, the target tree algorithm was proposed, substituting a parking goal in RRT with a set (target tree) of backward parking paths.
However, it consists of circular and straight paths, and an autonomous vehicle cannot park accurately because of curvature-discontinuity.
Moreover, the planning time increases in complex environments; backward paths can be blocked by obstacles.
Therefore, this paper introduces the continuous-curvature target tree algorithm for complex parking environments.
First, a target tree includes clothoid paths to address such curvature-discontinuity.
Second, to reduce the planning time further, a cost function is defined to construct a target tree that considers obstacles.
Integrated with optimal-variant RRT and searching for the shortest path among the reached backward paths, the proposed algorithm obtains a near-optimal path as the sampling time increases. 
Experiment results in real environments show that the vehicle more accurately parks, and continuous-curvature paths are obtained more quickly and with higher success rates than those acquired with other sampling-based algorithms.
\end{abstract}

\begin{IEEEkeywords}
Path Planning, Autonomous Parking, Target Tree algorithm, Rapidly-exploring Random Tree.
\end{IEEEkeywords}

\IEEEpeerreviewmaketitle

\section{Introduction}\label{sec:intro}
\IEEEPARstart{P}{ath} planning is a key component in autonomous parking tasks \cite{wada2003development, idris2009car}.
Path planning methods for parking need to satisfy the following conditions. 
First, a collision-free path should be planned considering various obstacles around the parking spot.
Second, the parking path should be obtained within a short planning time, even in complex parking situations.
Third, the path needs to be a continuous-curvature path for the autonomous vehicle to track and park accurately. 
In other words, the method needs to consider the vehicle's kinematic constraints, such as minimum turning radius and maximum steering velocity.

Various path planning methods take the abovementioned conditions into account, such as geometric, optimization-based, grid search-based, and sampling-based methods.
A geometric method plans the parking path in a short planning time by a combination of simple geometric curves that considers the vehicle's constraints \cite{inoue2004development, min2012design, kim2014auto, kim2020perpendicular, 7995715}. 
However, this approach may fail to find the path in complex parking environments where the road is narrow due to obstacles near the parking spot.
Optimization-based methods have been applied in various parking situations \cite{zips2016optimisation, moon2017real, li2019tractor}; they formulate the path planning problem as an optimization problem.
This optimization problem needs to be convexified to find the path within a short planning time.
A grid search-based method searches the parking path in a grid unit while considering the vehicle's constraints \cite{siedentop2015path, chen2015path, sedighi2019guided}.
This technique can certainly find the path and needs a short planning time in simple parking situations.
However, the path quality depends on the grid resolution.
The planning time can increase with a higher grid resolution.

Rapidly-exploring random tree (RRT) \cite{lavalle1998rapidly}, a typical sampling-based method, was studied for parking in \cite{han2011unified, shin2016desired, wang2017two, manav2021novel}. 
RRT is guaranteed to find a collision-free path, if one exists, and the path does not depend on the discretization.
Compared with the optimization- and grid search-based methods, RRT can plan the path within a shorter planning time when the road is narrow and obstacles surround the parking spot.
Nevertheless, the planned path may vary in accordance with random samples. 
Furthermore, the planning time can be further increased by the low probability that the random sample is connected to the tree without collisions at a narrow parking spot and in a narrow road.

The target tree algorithm \cite{feng2018model} was proposed to find the parking path in a shorter planning time than RRT path planning algorithm.
The target tree algorithm substitutes a parking goal in RRT with a set (target tree) of backward parking paths, and each backward path consists of candidate goals.
If one of the candidate goals is reached by RRT, then the parking path is found.
The target tree algorithm can reduce the planning time by building the backward parking paths in advance such that narrow regions need not be searched by RRT.
However, the autonomous vehicle cannot easily track the path and park accurately due to the curvature-discontinuity between the circular and straight paths in the target tree.
A parking path with such curvature-discontinuity does not take into account the vehicle's steering velocity.
Moreover, the target tree should be constructed considering the obstacles in complex parking environments to reduce the planning time.
This is because the planning time can be further increased or decreased according to the narrow regions covered by the target tree.

To address these limitations, this paper presents the continuous-curvature target tree algorithm for complex autonomous parking situations.
First, the proposed target tree algorithm builds a continuous-curvature target tree, which considers the vehicle's maximum steering velocity.
This target tree allows the vehicle to track the path and park more accurately than it can with using the original target tree algorithm.
Second, a cost function is proposed to construct a target tree that considers obstacles in complex parking situations.
The target tree that can minimize the area to be searched by RRT is identified by using this cost function, and the planning time is reduced.
The proposed algorithm is integrated with optimal-variant RRT (RRT*) and searches for the minimum-length path among the randomly reached candidate goals, thereby obtaining a shorter parking path within the given sampling time.

The rest of this paper is organized as follows.
In Section \ref{sec:related_works}, other path planning methods for parking are introduced. 
In Section \ref{sec:background}, the background of the target tree algorithm \cite{feng2018model} and its limitations are discussed.
Section \ref{sec:method} details the proposed path planning algorithm in three parts: i) continuous-curvature target tree, ii) cost function for reducing the planning time, and iii) minimum-length path selection.
The experimental results are presented in Section \ref{sec:exp}.
Section \ref{sec:conclusion} concludes this study.

\section{Related Works}\label{sec:related_works}
The path planning methods for autonomous parking can be categorized into four groups \cite{banzhaf2017future}: geometric, optimization-based, grid search-based, and sampling-based methods.

Geometric methods are presented in \cite{inoue2004development, min2012design, kim2014auto, kim2020perpendicular, 7995715}.
These methods plan a path on which the vehicle drives out of the parking spot with a set of geometric curves, such as a circular arc or a straight line, and find a path connecting this path to the vehicle's initial pose using other geometric curves. 
Geometric methods are simple and can plan the parking path with a shorter planning time than can optimization-, grid search-, and sampling-based methods.
However, geometric approaches may fail to find the path in complex parking situations where the road is narrow due to obstacles near the parking spot.
Moreover, their use can be limited due to the assumption that the path contains one forward/backward direction switch \cite{inoue2004development, kim2014auto, kim2020perpendicular}.

Optimization-based methods formulate parking path planning as an optimal control problem (OCP) \cite{zips2016optimisation, moon2017real, li2019tractor}, and solve it by numerical optimization. 
These methods can consider desired parking behaviors as the objective function, such as minimization of the number of forward/backward direction switches, minimization of path length, or maximization of the obstacle clearance of the path.
Any vehicle's kinematic constraints can be considered by the equality/inequality constraints. 
However, the optimization problem needs to be convexified to find the path within a relatively short time.
Approximating complex parking situations to be convexified can be difficult, and an inaccurate path can be obtained.

Grid search-based methods were applied for parking path planning in \cite{siedentop2015path, chen2015path, sedighi2019guided}. 
A grid search-based path planning approach, such as Hybrid-A* \cite{dolgov2008practical}, discretizes the configuration space as a set of grids and incrementally searches these grids to find the path. 
This method is guaranteed to find a collision-free path if the path exists, and this parking path can be obtained within a short planning time in simple parking situations.
However, high-resolution grids can be required for parking paths when parking needs several forward/backward direction switches.
These high-resolution grids can exponentially increase the planning time.

A sampling-based method, typically RRT \cite{han2011unified, shin2016desired, wang2017two, manav2021novel}, finds the path by incrementally searching the configuration space through random samples. 
It does not depend on the discretization of the grids, and the path can be obtained within a shorter planning time in complex parking situations than can optimization- and grid search-based methods.
However, as mentioned in Section \ref{sec:intro}, the parking path may vary, and the planning time can be increased by narrow regions.
RRT* \cite{karaman2010incremental} can keep the parking path from varying through a tree-rewiring step, but this step requires further planning time.
There are bidirectional approaches \cite{kuffner2000rrt, jordan2013optimal, klemm2015rrt, jhang2020autonomous} that reduce the planning time by growing another tree from the parking spot and connecting the two trees.
These approaches can search for a narrow parking spot in advance. However, the planning time can increase if the two trees grow toward directions that are difficult to connect.

\section{Background: \\ Target Tree Algorithm and its Limitations}\label{sec:background}
\begin{figure}[t!] 
\centering
\subfigure[]{\includegraphics[width=0.980\linewidth]{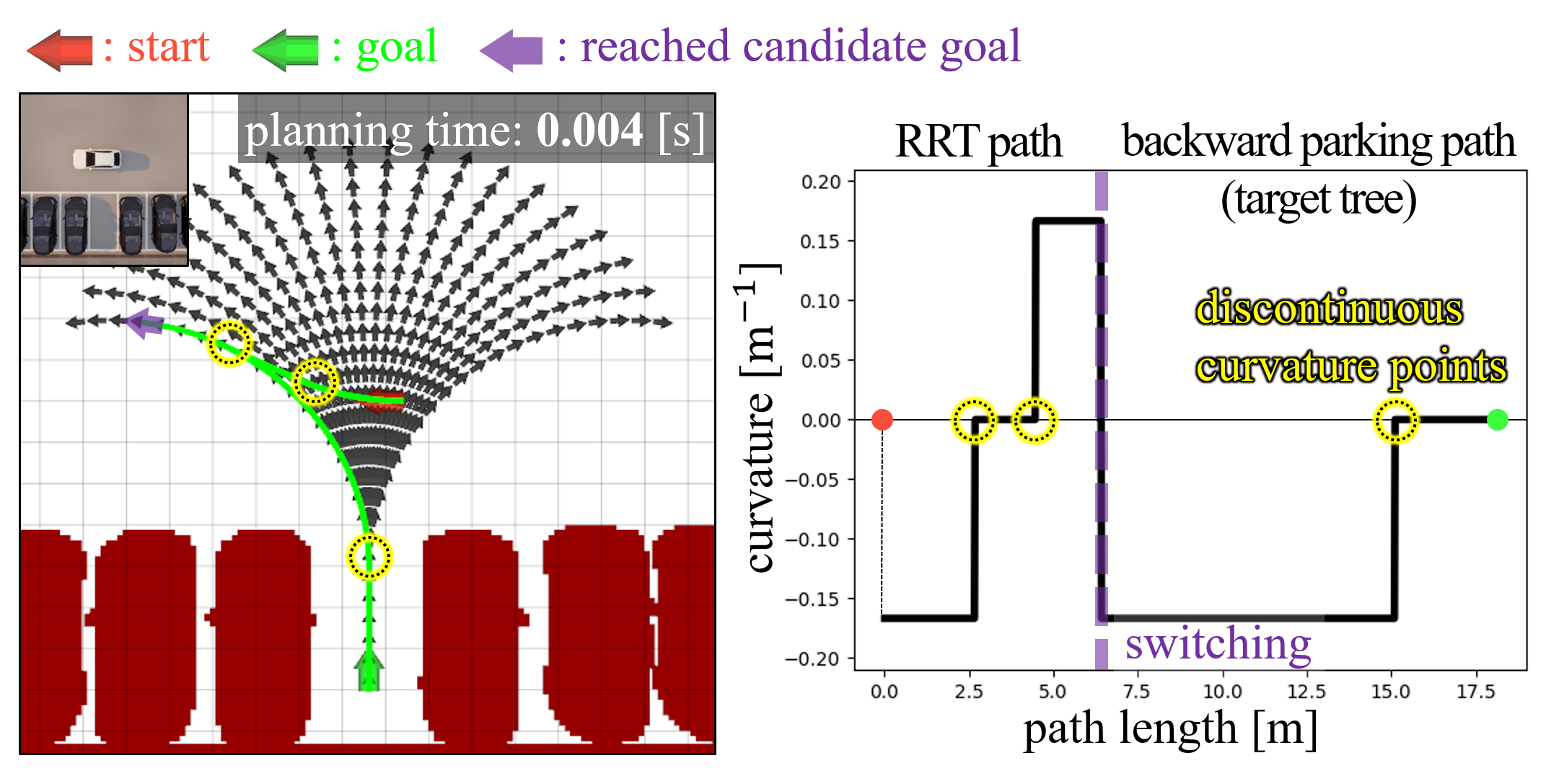}}
\subfigure[]{\includegraphics[width=0.455\linewidth]{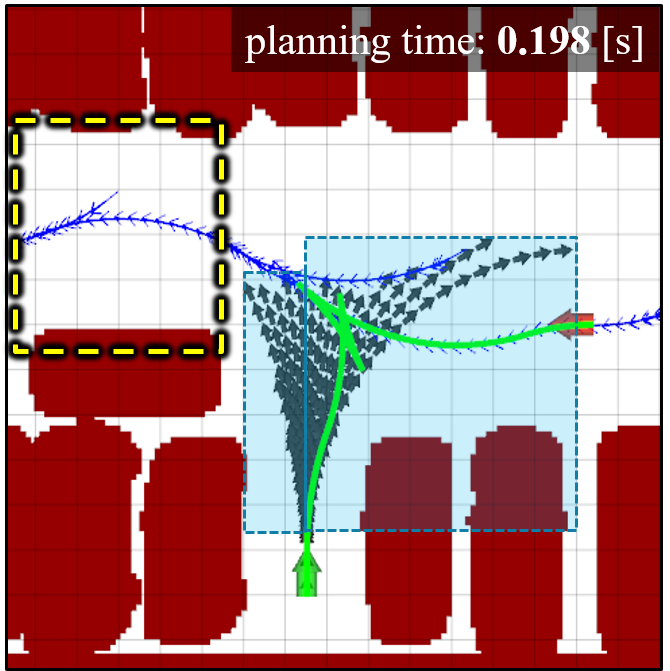}}
\hspace{0.14in}
\subfigure[]{\includegraphics[width=0.455\linewidth]{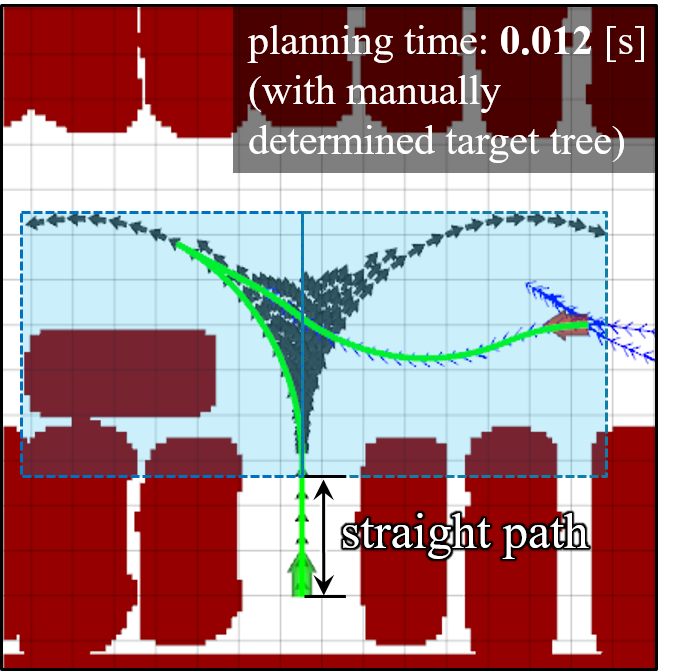}}
\caption{Path planning for parking with the target tree algorithm and RRT. 
The set of black arrows is the target tree, and the set of blue arrows is the RRT tree. 
The green trajectory is the planned path.
(a) The path includes discontinuous-curvature points (yellow dots).
(b) The planning time increases when the target tree does not cover the narrow region (yellow rectangle). 
(c) It can be reduced by the target tree covering larger areas that include this narrow region.
} 
\label{fig:1}
\end{figure}

The target tree algorithm was proposed in \cite{feng2018model} for parking path planning with RRT to reduce planning time. 
The target tree algorithm is detailed in Algorithm \ref{alg:1}.
First, the target tree is initialized (line 1 in Algorithm \ref{alg:1}), as shown in Fig. \ref{fig:1}(a).
The algorithm builds a set of backward parking paths on which the parked vehicle drives out of the parking spot without collisions, and this set is defined as the target tree ($T_{target}$).
Each backward parking path, as a branch of the target tree, consists of a straight path and a circular path.
The straight path is determined by the vehicle driving straight out of the parking spot.
The circular path is built from the end of this straight path.
Each backward parking path is built by changing the circular path's curvature within \{$-\kappa_{\text{max}}$, ..., $\kappa_{\text{max}}$\}.
The backward parking path of the target tree is discretized as a set of poses (black arrows in Fig. \ref{fig:1}), and each pose becomes a candidate goal to be reached by the RRT tree ($T$).
If one of these goals is reached, then the parking path is obtained (lines 9-11 in Algorithm \ref{alg:1}).
However, the target tree algorithm has two limitations.

\begin{algorithm}
\label{alg:1}
\caption{Target tree algorithm \cite{feng2018model}
\\ {\color{blue}* The blue font details the proposed algorithm.}
}
 \begin{algorithmic}[1]
 \renewcommand{\algorithmicrequire}{\textbf{Input:}}
 \renewcommand{\algorithmicensure}{\textbf{Output:}}
 \REQUIRE $q_{init}$, $q_{goal}$
 \ENSURE  $path$
  \STATE $T_{target} \leftarrow$ InitializeTargetTree($q_{goal}$); \\ 
  {\color{blue} // The continuous-curvature target tree is initialized \\ by using a cost function (Algorithm \ref{alg:2}, in Section \ref{subsec:cost_function})}
  \STATE $T \leftarrow$ InitializeTree($q_{init}$);
\WHILE{$t < t_{max}$}
  \STATE $q_{rand} \leftarrow$ Sample($T_{target}$, $\tau$);
  \STATE $q_{nearest} \leftarrow$ Nearest($q_{rand}, T$);
  \STATE $q_{new} \leftarrow$ Steer($q_{nearest}, q_{rand}$);
   \IF{ObstacleFree($q_{nearest}$, $q_{new}$)}
    \STATE $T \leftarrow$ InsertNode($q_{new}$);
        \IF{ $q_{new} \in T_{target}$ }
            \STATE $path \leftarrow$ ExtractPath($T, q_{new}, T_{target}$);\\
            \STATE \textbf{break}; \\ 
            {\color{blue} // Lines 10 and 11 are replaced by Eqs. \ref{eq:5} and \ref{eq:6} \\ to find a shorter parking path (in Section \ref{subsec:minimum_length_path_selection})}
        \ENDIF
    \ENDIF
\ENDWHILE
 \RETURN $path$ 
 \end{algorithmic}
\end{algorithm}

First, the target tree includes a discontinuous-curvature point (yellow dot in Fig. \ref{fig:1}(a)) between the straight and circular paths.
This point can increase the probability of collision while the autonomous vehicle tracks the path and parking alignment error when the vehicle is parked.
This is because when the vehicle tracks from the circular path to the straight path, it needs to instantly change the steering angle from the maximum steering to zero at this discontinuous point.
A real vehicle cannot achieve this due to its maximum steering velocity.
Thus, the vehicle may deviate from the path and cannot be parked accurately at the parking spot.
This problem may be addressed by an additional path-smoothing step that converts this discontinuous-curvature path to a continuous-curvature one.
However, it requires additional computational time because the vehicle's kinematic constraints and collisions will be re-considered at a narrow parking spot.
This curvature-discontinuity problem is addressed by the proposed continuous-curvature target tree in Section \ref{subsec:continuous_curvature_target_tree}.

Second, a target tree that does not consider obstacles can increase the planning time in complex parking environments.
This is because, the area of the road covered by the target tree (cyan rectangles in Fig. \ref{fig:1}) varies depending on the environment, and the planning time can be further increased or decreased.
For example, Fig. \ref{fig:1}(b) shows an area that the target tree cannot cover due to obstacles (yellow rectangle).
Accordingly, this area needs to be searched by the RRT tree, and the planning time may increase.
In Fig. \ref{fig:1}(c), if the length of the straight path in the target tree is increased to cover a larger area of the road, then the planning time can be reduced.
However, the target tree needs to be constructed manually for each parking situation.
The proposed cost function deals with this problem by calculating the cost of the target tree and identifying a proper target tree for each parking situation. This process is detailed in Section \ref{subsec:cost_function}.

The proposed target tree algorithm is integrated with RRT* instead of RRT and searches for a shorter path (i.e., minimum-length path selection).
This integration allows the target tree algorithm to obtain the minimum-length parking path (near-optimal path) within the given sampling time ($t_{max}$), even after the first parking path is found (lines 10 and 11 in Algorithm \ref{alg:1}).
This procedure is described in Section \ref{subsec:minimum_length_path_selection}.

\section{Continuous-Curvature Target Tree Algorithm for Complex Parking Environments}\label{sec:method}
\subsection{Continuous-Curvature Target Tree}\label{subsec:continuous_curvature_target_tree}
The proposed continuous-curvature target tree addresses the curvature-discontinuity of the original target tree \cite{feng2018model} by using a clothoid path.
A clothoid path was used to find a continuous-curvature path by addressing the curvature-discontinuity between straight lines and circular arcs in \cite{fraichard2004reeds, banzhaf2017hybrid}.
Motivated by these works, the continuous-curvature target tree in the present work is defined as a set of backward parking paths with continuous curvature by including a clothoid path between the straight and circular paths.
In contrast to the original target tree, the continuous-curvature target tree considers not only the vehicle's minimum turning radius but also its steering velocity.

The vehicle model for the continuous-curvature target tree is defined as a kinematic bicycle model \cite{rajamani2011vehicle}, and additionally considers the curvature derivative of the path.
The vehicle model is defined as 
\begin{align}
    \begin{pmatrix}x' \\ y' \\ \theta' \\ \kappa' \end{pmatrix} =
    \begin{pmatrix}cos(\theta) \\ sin(\theta) \\ \dfrac{tan(\delta)}{L} \\ 0 \end{pmatrix} d
    + \begin{pmatrix}0 \\ 0 \\ 0 \\ \sigma \end{pmatrix},
\end{align}
where $(x, y)$ is the position of the center of the rear axle.
$\theta$ means the orientation of the vehicle.
$L$ is the wheelbase of the vehicle, and $\delta$ is the vehicle's steering angle.
$d$ is the forward or backward direction of the vehicle.
$(\bullet')$ means the derivative with respect to the path length.
$\kappa$ ($=\theta'$) is the curvature of the path.
$\sigma$ means the sharpness, the curvature derivative.
The sharpness is related to $\delta'$, the vehicle's steering velocity of the front wheels ($\kappa'=\dfrac{\delta'}{L cos^2(\delta)}$).
The curvature of the path $\kappa$ and its derivative $\sigma$ are constrained by $|\kappa| \leq \kappa_{\text{max}}$ and $|\sigma| \leq \sigma_{\text{max}}$.

\begin{figure}[t!] 
\centering
\subfigure[]{\includegraphics[width=0.99\linewidth]{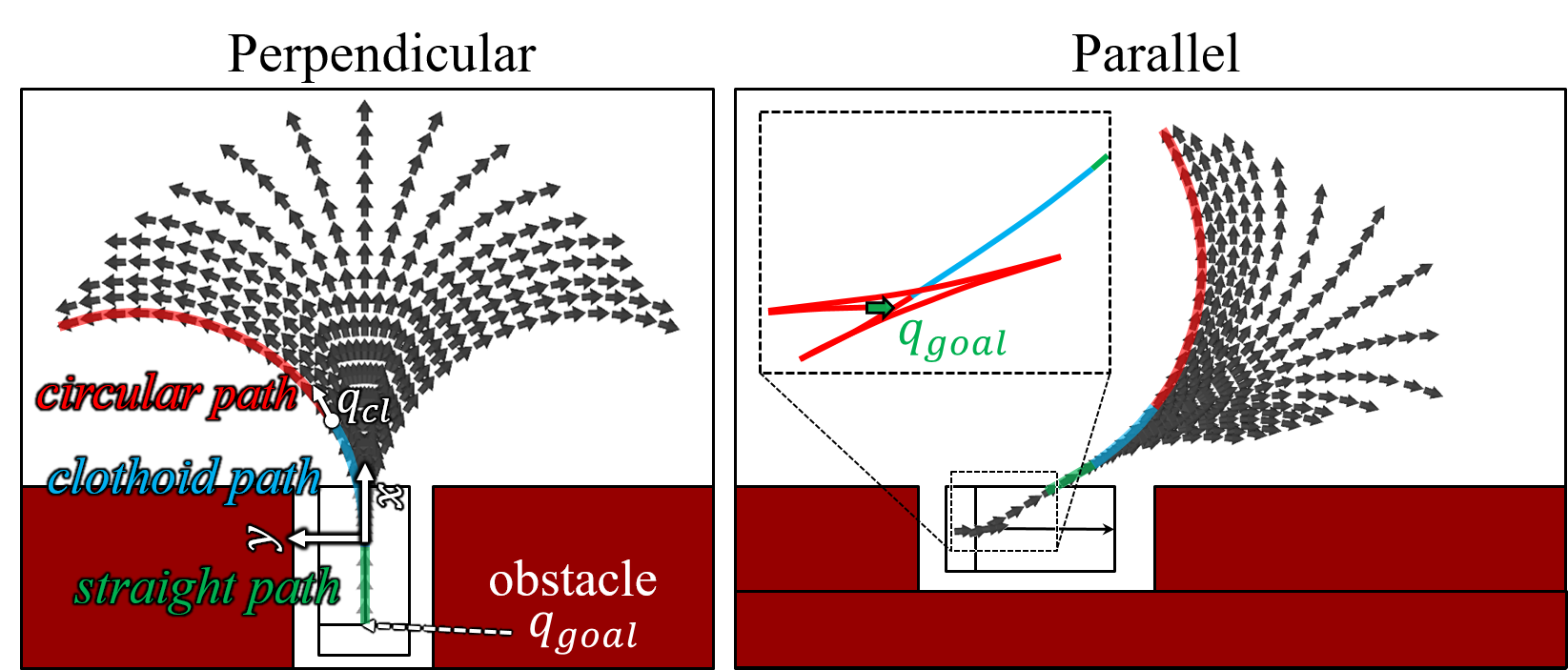}}
\subfigure[]{\includegraphics[width=0.99\linewidth]{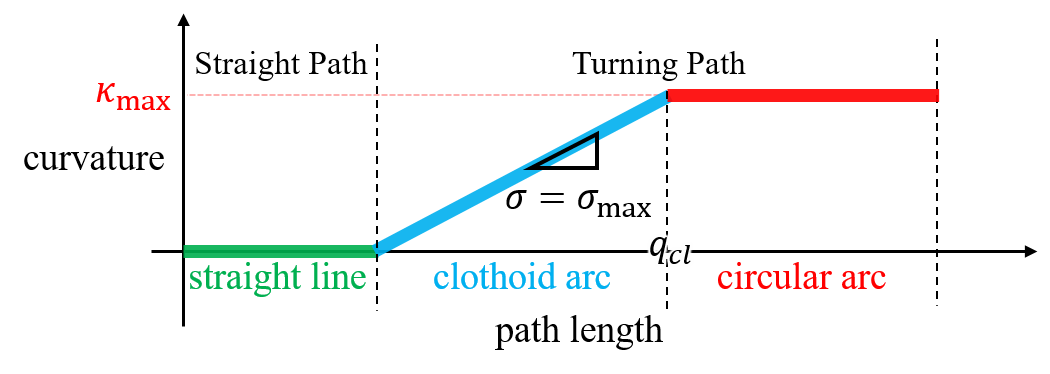}}
\subfigure[]{\includegraphics[width=0.99\linewidth]{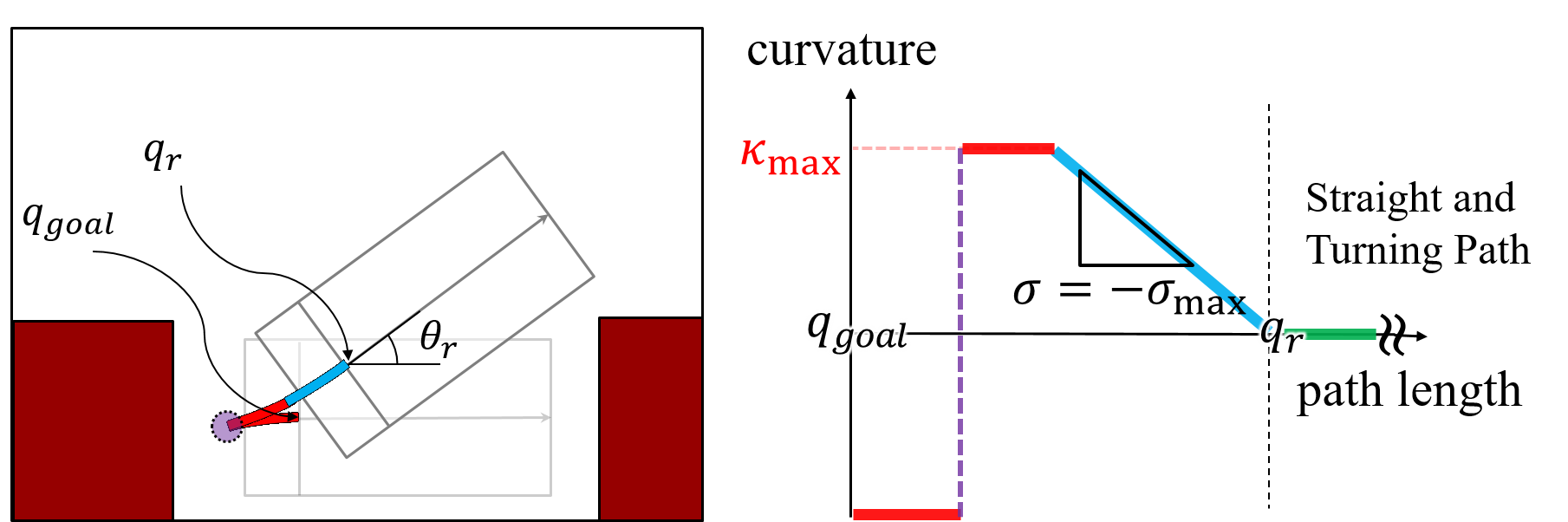}}
\caption{Descriptions of the continuous-curvature target tree. 
(a) Continuous-curvature target tree for perpendicular and parallel parking. 
The target tree is represented as the set of the black arrows.
The green, cyan, and red solid lines are the straight, clothoid, and circular paths, respectively. 
(b) The plot represents the leftmost path of the target tree for perpendicular parking. 
(c) A set of circular paths, and a clothoid path are added for parallel parking. 
}
\label{fig:2}
\end{figure}

The continuous-curvature target tree is divided into perpendicular and parallel parking cases (see Fig. \ref{fig:2}).
For perpendicular parking, the continuous-curvature target tree consists of straight, clothoid, and circular paths (left part of Fig. \ref{fig:2}(a)).
For parallel parking, the continuous-curvature target tree is designed by adding a set of backward and forward circular paths, and a clothoid path (right part of Fig. \ref{fig:2}(a)).
The vehicle can be parked by aligning it to the parking spot by moving forward and backward via these additional paths.

\begin{figure*}[t!] 
\includegraphics[width=7.1in]{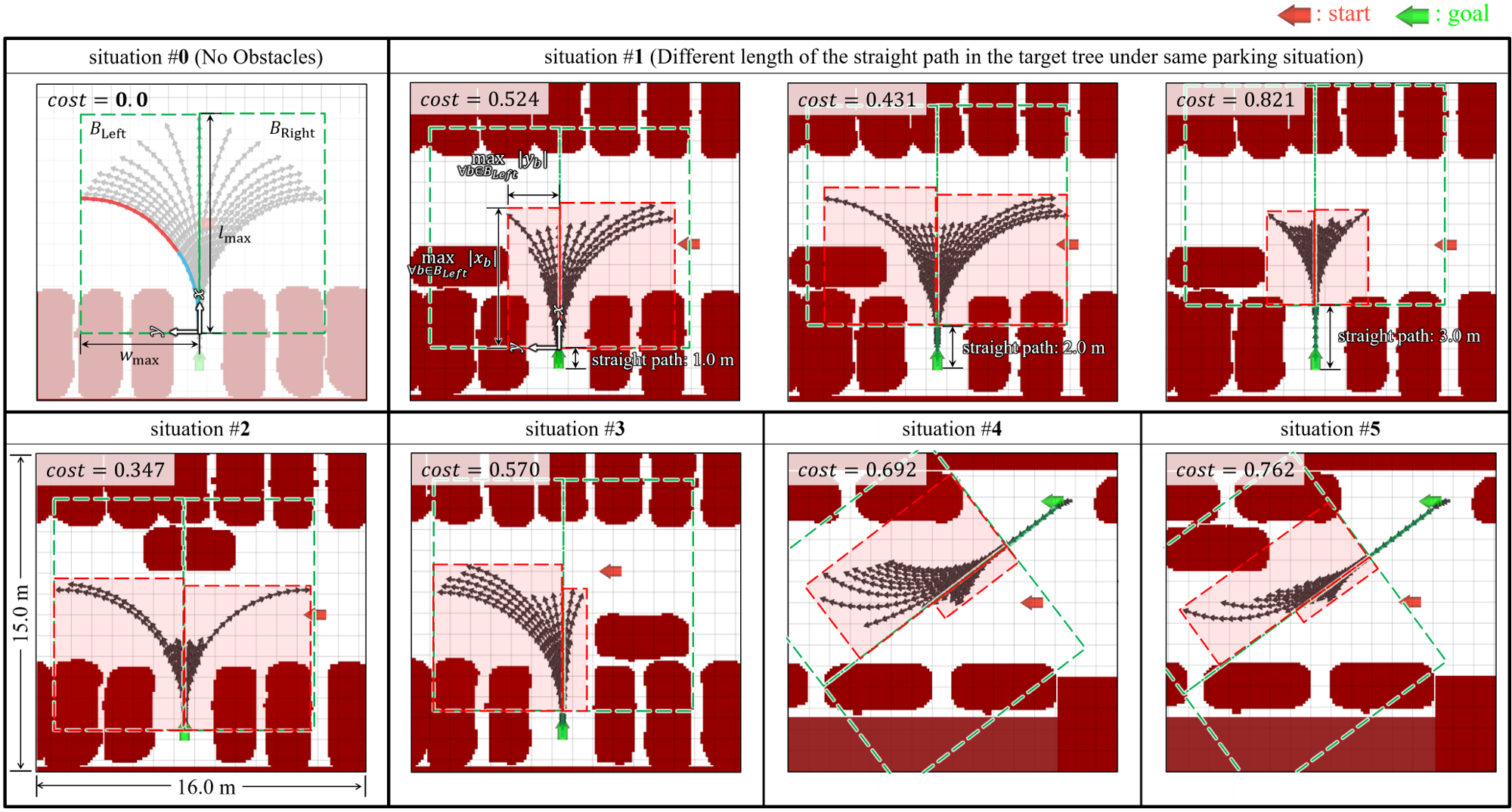}
\centering 
\caption{Description of the cost for each target tree. 
The light gray grid measures 1 m $\times$ 1 m.
$B_{Left}$ and $B_{Right}$ are the left and right branches, respectively, of the target tree with respect to the $x$-axis.
$l_{\text{max}}$ and $w_{\text{max}}$ are the maximum length and width of the rectangle, respectively, when there are no obstacles (situation \#\textbf{0}). 
The green rectangles denote the area that the target tree covers when there are no obstacles.
The red shaded area is $\Sigma_k A_k$, which the target tree covers when there are obstacles. 
}
\label{fig:3}
\end{figure*}

The clothoid path is a path of which curvature is changed linearly by the path length. The curvature, $\kappa$, is described as 
\begin{align}
    \theta'(s) = \kappa(s) = \sigma s,
\end{align}
where $s$ is the path length.
In Fig. \ref{fig:2}(b), the curvature of the clothoid path (cyan line) is initially zero.
It is changed linearly from zero to the maximum curvature, $\kappa_{\text{max}}$, by the sharpness, $\sigma$, as the slope in the path length and curvature plot.
After the clothoid path, the vehicle's configuration, $q_{cl}$, can be derived with respect to the $xy$-axis in Fig. \ref{fig:2}(a), given as
\begin{align}
    q_{cl} = \begin{pmatrix}
         x_{cl} \\  y_{cl} \\  \theta_{cl} \\ \kappa_{cl}
    \end{pmatrix}
    =
    \begin{pmatrix}\sqrt{\pi/\sigma}C_f(\sqrt{\kappa_{\text{max}}^2/(\pi\sigma)}) \\ \sqrt{\pi/\sigma}S_f(\sqrt{\kappa_{\text{max}}^2/(\pi\sigma)}) \\ \kappa_{\text{max}}^2 / (2\sigma) \\ \kappa_{\text{max}} \end{pmatrix},
\end{align}
where $(x_{cl}, y_{cl}, \theta_{cl})$ is the vehicle's pose after the curvature changes from zero to $\kappa_{\text {max}}$ through the clothoid path. $C_f(s)$ and $S_f(s)$ are Fresnel Integrals \cite{mccrae2009sketching} defined as $C_f(s)=\int_0^scos(\frac{\pi}{2}u^2)du$ and $S_f(s)=\int_0^ssin(\frac{\pi}{2}u^2)du$.

The sharpness, $\sigma$, is changed within $\{-\sigma_{\text{max}}, ..., \sigma_{\text{max}}\}$ to build the branches of the continuous-curvature target tree.
For example, the branch of the target tree in the left part of Fig. \ref{fig:2}(a) (the green, cyan and red lines) is constructed by setting the sharpness as $\sigma_{\text{max}}$.
As the sharpness, $\sigma$, decreases from $\sigma_{\text{max}}$ to $-\sigma_{\text{max}}$, the other branches are formed.
At each branch, when the clothoid path's curvature becomes the maximum curvature, $\kappa_{\text{max}}$, the circular path whose radius is $\kappa_{\text{max}}^{-1}$ (red line) is added to the end of the clothoid path (see Fig. \ref{fig:2}(b)).
If the curvature does not reach the maximum curvature due to collisions, the branch will consist of straight and clothoid paths.

For parallel parking, a set of circular paths, and a clothoid path are added to the straight and turning paths.
This set of circular paths was proposed in \cite{vorobieva2014automatic} as a method by which a vehicle drives out of a parking spot in parallel parking.
The continuous-curvature target tree for parallel parking is described in Fig. \ref{fig:2}(c).
The backward circular path is built by the vehicle driving backward from the parking spot, $q_{goal}$, to approach the rear obstacle without collisions.
The vehicle's direction is switched to forward, and the forward circular path is added.
If the vehicle can drive out of the parking spot with these circular paths, the clothoid path whose curvature is changed from $\kappa_{\text{max}}$ to zero is added to the end of this forward circular path.
Otherwise, another set of backward and forward circular paths is added, and the above steps is repeated.
For example, in the right part of Fig. \ref{fig:2}(a), two sets of backward and forward circular paths, and the clothoid path are built.
The number of sets of circular paths is increased as the dimensions of the parking spot are decreased \cite{vorobieva2014automatic}.

\subsection{Cost Function for Reducing Planning Time in Complex Parking Environments}\label{subsec:cost_function}
The cost function is proposed to determine the target tree that can reduce the planning time.
The target tree that covers a larger area of the road for a parking situation is identified, by calculating the cost of the target tree.
This is because the area to be searched by RRT can be reduced if the road could be covered widely by the target tree.
In this regard, the planning time can be reduced further if the target tree covers a larger area of the road.
The larger the area that the target tree covers, the lower the cost of the target tree.

Examples for calculating the cost of the target tree are shown in Fig. \ref{fig:3}.
The cost function uses the area covered by the turning path of the target tree.
The area is divided into two portions with respect to the straight path of the target tree.
It is represented as the red rectangles.
Each portion is calculated by the length and width of each rectangle.
The length (width) is defined as the maximum distance of the turning path along the $x$-axis ($y$-axis).
The cost function is defined as 
\begin{align}
    \begin{aligned}[b]
    cost &= 1 - (\Sigma_k A_k) / (2l_{\text{max}}w_{\text{max}}),\\
    A_k &= \max_{\forall b\in B_k}\abs{x_b} \times \max_{\forall b\in B_k}\abs{y_b},
    \end{aligned}
    \label{eq:4}
\end{align}
where $x_b$ and $y_b$ are the coordinates at the end of each branch $b$ in the target tree. 
$B_k$ means the set of the branches at \{Left, Right\} in the target tree in Fig. \ref{fig:3}.
$\max_{\forall b\in B_k}\abs{x_b}$ and $\max_{\forall b\in B_k}\abs{y_b}$ are the length and width, respectively.
Thus, $A_k$ means the area of each red rectangle, and $\Sigma_k A_k$ means the area covered by the turning paths of the target tree (see Fig. \ref{fig:3}).
$l_{\text{max}}$ and $w_{\text{max}}$ are the length and width, respectively, when there are no obstacles blocking the turning paths of the target tree.
In the case of Fig. \ref{fig:3} (top-left), $l_{\text{max}}$ becomes the length of the turning path.
$w_{\text{max}}$ is calculated by the distance of the turning path along the $y$-axis, in the leftmost branch.
Hence, $2l_{\text{max}}w_{\text{max}}$ in (\ref{eq:4}) denotes the area covered by the target tree when there are no obstacles, and it is depicted as the green rectangles. 
$(\Sigma_k A_k) / (2l_{\text{max}}w_{\text{max}})$ in (\ref{eq:4}) becomes the ratio of the area covered by the target tree in the parking situation to the area covered by the target tree when there are no obstacles around the turning paths.

The proposed target tree algorithm initializes the target tree using (\ref{eq:4}), as detailed in Algorithm \ref{alg:2}.
\begin{algorithm} 
\caption{{\bf InitializeTargetTree($q_{goal}$)}}
 \begin{algorithmic}[1]
 \renewcommand{\algorithmicrequire}{\textbf{Input:}}
 \renewcommand{\algorithmicensure}{\textbf{Output:}}
 \REQUIRE $q_{goal}$
 \ENSURE  $T_{target}$
  \STATE {\bf T} $= \emptyset$;
  \FOR {\texttt{$l=0, \alpha, 2\alpha, ..., l_{parking}$}}
  \STATE $T_{target} \leftarrow$ ContinuousCurvatureTargetTree($q_{goal}$, $l$);
  \STATE $cost \leftarrow $ CalculateCost($T_{target}$);
  \STATE {\bf T} $\leftarrow$ pushback(($T_{target}$, $cost$));
  \ENDFOR
  \RETURN GetMinimumCostTree({\bf T});
 \end{algorithmic}
 \label{alg:2}
\end{algorithm}
For determining the target tree that considers obstacles, continuous-curvature target trees are built by changing the length of the straight path, $l$, from zero to the length of the parking spot, $l_{parking}$, by intervals of $\alpha$ (lines 2-6 in Algorithm \ref{alg:2}).
The $cost$ of each target tree is calculated by the proposed cost function in (\ref{eq:4}) (line 4 in Algorithm \ref{alg:2}).
For example, in parking situation \#\textbf{1} in Fig. \ref{fig:3}, the $cost$ of the target tree is calculated with different lengths of the straight path, $l$.
The continuous-curvature target tree, $T_{target}$, with minimum $cost$ is selected among those target trees (line 7 in Algorithm \ref{alg:2}).

\subsection{Minimum-Length Path Selection for Finding Shorter Parking Path}\label{subsec:minimum_length_path_selection}
The proposed target tree algorithm finds a shorter parking path by integrating with RRT* and executing the minimum-length path selection step.
RRT* rewires a tree (denoted as $T$ in Algorithm \ref{alg:1}) to reduce the length of the RRT* path that connects $q_{init}$ to a reached candidate goal, $q_{new}$.
The proposed path selection step searches for a shorter parking path among the randomly reached candidate goals of the target tree within the sampling time ($t_{max}$ in Algorithm \ref{alg:1}).

The minimum-length path selection step replaces lines 10 and 11 in Algorithm \ref{alg:1}.
First, several candidate goals of the target tree reached by the RRT* tree, $T$, are stored within the sampling time.
This is because, even after the first parking path is obtained, RRT* paths that reach other candidate goals may be found during the additional time for a tree-rewiring step.
Lines 10 and 11 in Algorithm \ref{alg:1} are replaced by,
\begin{align}
    \begin{aligned}[b]
    Q_{soln} \leftarrow Q_{soln} \cup \{q_{new}\},
    \end{aligned}
    \label{eq:5}
\end{align}
where $q_{new}$ is the candidate goal in the target tree reached by the RRT* tree, $T$.
$Q_{soln}$ is a set of these goals. 
Next, the minimum-length path selection step is added as follows,
\begin{align}
    \begin{aligned}[b]
    path \leftarrow \text{\textbf{GetMinimumLenPath}(}T, Q_{soln}, T_{target}\text{)}.
    \end{aligned}
    \label{eq:6}
\end{align}
In the \textbf{GetMinimumLenPath} function, the shortest parking path is returned among a set of the reached candidate goals, $Q_{soln}$.
This function is executed when one of the candidate goals is reached or when the RRT* tree, $T$, is rewired.
This minimum-length path selection step allows the target tree algorithm to find a shorter parking path as the sampling time increases.
It is similar to the RRT* path planning algorithm, which finds a shorter path by adding a tree-rewiring step to RRT.
The path selection step can also be used when the cost function for tree rewiring considers not only the path length but also other costs such as the number of forward/backward direction switches, and obstacle clearance.

\section{Experiments and Discussions}\label{sec:exp}
\subsection{Experimental Setup}\label{subsec:exp_setup}
\begin{figure}[t!] 
\centering
\includegraphics[width=2.90in]{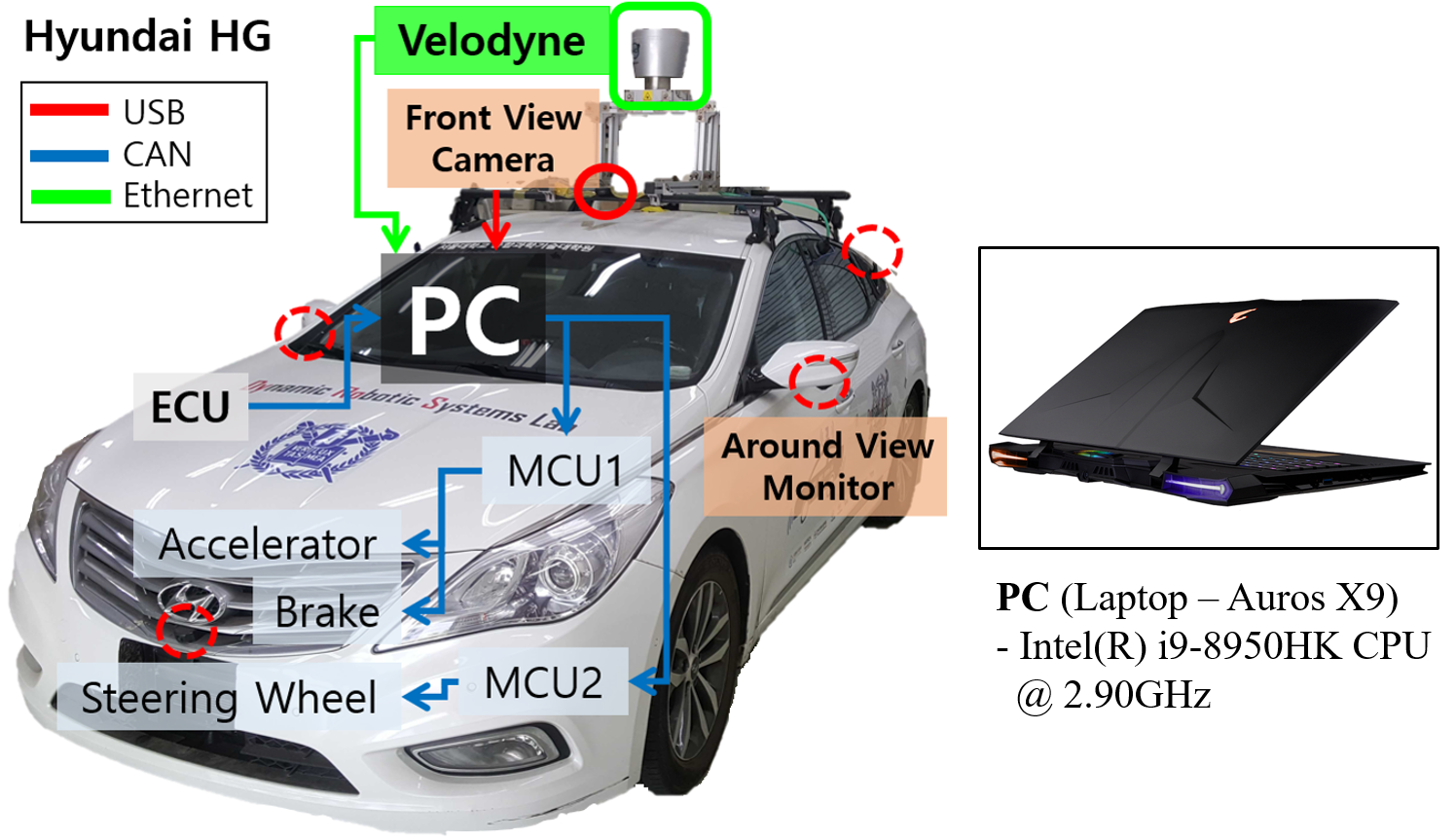}
\caption{Hardware configuration of the autonomous vehicle. The autonomous vehicle belongs to the Dynamic Robotic Systems (DYROS) laboratory at Seoul National University.
}
\label{picture}
\label{fig:4}
\end{figure}

The proposed algorithm was tested with an autonomous vehicle in real parking environments.
The hardware configuration of the autonomous vehicle is shown in Fig. \ref{fig:4}.
The proposed target tree algorithm was implemented with Open Motion Planning Library (OMPL) \cite{sucan2012open}.
The vehicle's dimensions and path planning parameters are described in Table \ref{table:1}.

\begin{table}[h!]
\caption{Vehicle dimensions and planning parameters for experiments}
\renewcommand{\arraystretch}{1.0}
\centering
\begin{tabular}{| l | l |} 
 \hline\hline
    Parameter & Value \\ 
 \hline\hline
    vehicle length & 4.910 m  \\ 
 \hline
    vehicle width & 1.860 m  \\
 \hline
     wheel base ($L$) & 2.845 m  \\
 \hline
    max. curvature ($\kappa_{\text{max}}$) & 1/6.000 m\textsuperscript{-1}  \\
 \hline
    max. sharpness ($\sigma_{\text{max}}$) & 0.200 m\textsuperscript{-2}  \\
 \hline\hline
\end{tabular}
\label{table:1}
\end{table}

The hybrid curvature (HC) tree-extension function \cite{banzhaf2017hybrid} was used when the proposed target tree algorithm was integrated with RRT* for planning the continuous-curvature parking path.
A kanayama controller \cite{kanayama1990stable} was used for tracking the parking path.
The controller received the vehicle's position and orientation relative to the path and curvature of the path as inputs, and the vehicle's input steering angle was calculated.
The vehicle's maximum velocity was set to 4 km/h and 2 km/h when tracking the RRT path and the backward parking path of the target tree, respectively.
The vehicle's velocity was decreased in an inversely proportional manner to the path's curvature.
At the forward/backward direction switch, the vehicle stops for 3 s to change the steering angle.
The LeGO-LOAM algorithm \cite{shan2018lego} was used to obtain the vehicle's position and orientation relative to the path.

\subsubsection{Experiments for continuous-curvature target tree} \label{subsubsec:exp1}
In the first experiment, the effectiveness of the continuous-curvature target tree was evaluated.
The continuous-curvature target tree algorithm with RRT* and HC tree-extension function was compared with i) the original target tree algorithm with RRT \cite{feng2018model} and ii) the original target tree algorithm with RRT* and HC tree-extension function.
The parking path was planned by each algorithm, and the autonomous vehicle tracked the path and parked in perpendicular and parallel parking situations. 
The first experiment was repeated five times for each algorithm, and the path tracking and parking alignment errors were measured.

\subsubsection{Experiments for cost function} \label{subsubsec:exp2}
In the second experiment, the effectiveness of the proposed cost function was evaluated.
The path planning time was measured when the parking path was planned with the target tree determined by the proposed cost function.
The parking situations included a parallel-parked vehicle near the parking spot, in which parking required considering a narrow region (the road width was reduced to about 3.5 m), i.e., a complex parking environment.
Continuous-curvature target trees with different $cost$s were built in each parking situation.
Each target tree was built by discretely changing the length of the straight path by the interval, $\alpha$ (Algorithm \ref{alg:2}). 
The interval, $\alpha$ was set to 0.2 m considering the road width and the dimensions of the parallel-parked vehicle.
The proposed algorithm (using the minimum-$cost$ target tree) was compared with the target tree algorithm with higher-$cost$s target tree.
Also, it was compared with other sampling-based algorithms for parking, which uses informed-RRT* \cite{kim2021comparative} and bidirectional-RRT* \cite{jhang2020autonomous}, respectively.
In \cite{kim2021comparative}, the RRT* tree was built from the parking spot, and informed sampling was applied to reduce the planing time. 
In \cite{jhang2020autonomous}, bidirectional tree growth was combined with RRT* to deal with a narrow parking spot.
The path was planned 100 times for each algorithm.
The path length, planning time, and planning success rate were measured.
The maximum sampling time in RRT* ($t_{max}$ in Algorithm \ref{alg:1}) was set to 3 s for considering the tree-rewiring step.
The path planning was deemed to have failed when the path was not found within the sampling time.
The cost function for tree rewiring considered the path length.

\begin{figure*}[t]
\centering
\includegraphics[width=7.1in]{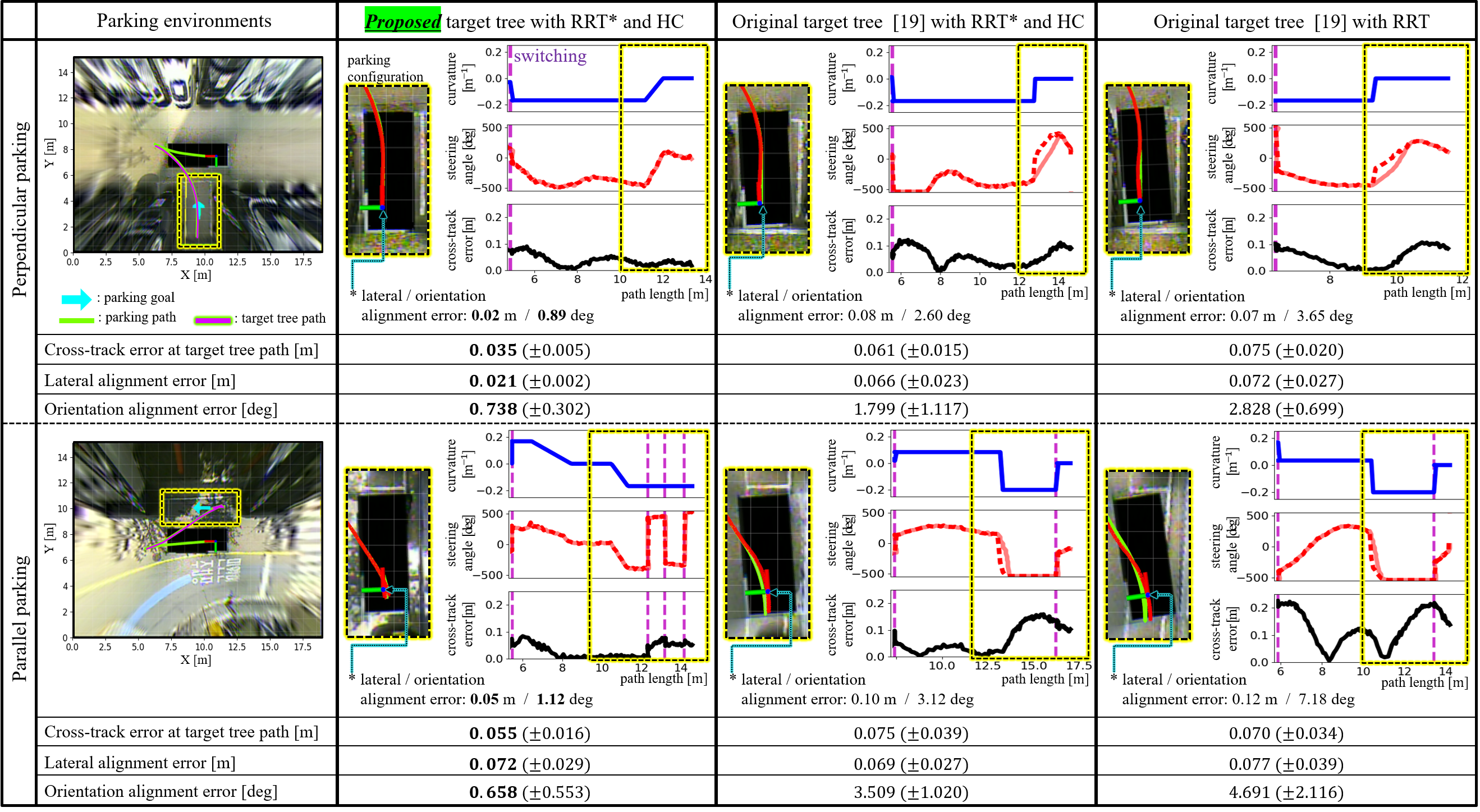}
\caption{
Path tracking results from five repetitions for each algorithm. 
In the tables, each error is described as 'mean $\pm$ standard deviation' in the five repetitions.
An example of one of the results in the five repetitions is shown above each table.
In a parking configuration, the red trajectory is the vehicle's trajectory.
In a steering angle plot, the red dashed line is the input steering angle by the controller, and the red solid line is the vehicle's actual steering angle.
}
\label{fig:5}
\end{figure*}

\subsection{Experimental Results}
\subsubsection{Analysis of results and discussion of continuous-curvature target tree experiments (Section \ref{subsubsec:exp1})}

The results of the first experiment where the vehicle tracks the path and parks are shown in Fig. \ref{fig:5}.
The parking result of the proposed algorithm is shown in a video\footnote{\url{https://youtu.be/8B7C-7Wg33w}}.
The cross-track error, which is the shortest distance between the vehicle's position and the closest point on the path while tracking the path, was calculated when the vehicle tracked the target tree path.
The lateral/orientation parking alignment errors were calculated by comparing the vehicle's pose and the parking goal when the vehicle was parked.
These parking alignment errors are criteria for determining whether a vehicle is parked properly.

As seen in Fig. \ref{fig:5}, the proposed target tree algorithm planned parking paths, where the vehicle tracked and parked with fewer tracking and parking errors.
The input steering angle was calculated considering the steering velocity, and the vehicle's steering angle (red solid line) was controlled with the input steering angle (red dashed line) without error.
When the vehicle was parked, in the proposed algorithm, all four wheels of the vehicle did not invade the parking line in contrast to the original algorithm \cite{feng2018model}.
In the perpendicular parking case, the lateral and orientation alignment errors were reduced by more than half compared with the case of the original target tree.
In the parallel parking case, the lateral alignment error was not significantly reduced because the vehicle moved forward and backward several times near the parking spot.
Nevertheless, these direction switches reduced the orientation alignment error by one-third compared with the tracking with the original target tree.

\subsubsection{Analysis of results and discussion of cost function experiments (Section \ref{subsubsec:exp2})}

\begin{figure*}[t] 
\centering
\includegraphics[width=7.08in]{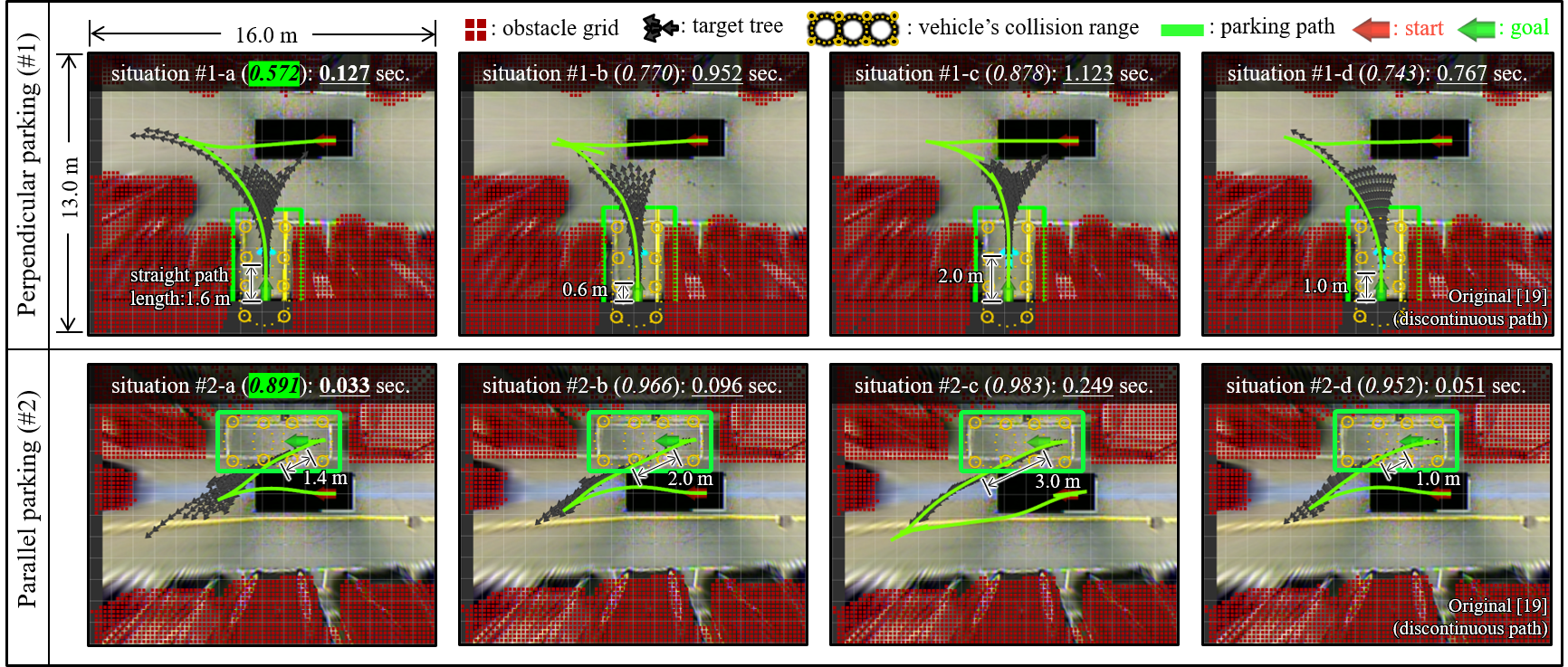}
\caption{
Parking situations for analyzing the relationship between the path planning performance and the cost of target tree under perpendicular (\#1) and parallel (\#2) parking.
Each result with a different target tree is represented as '\#number ($cost$): the mean time for the first solution by the target tree algorithm'.
The target tree in `\textbf{-a}' is the minimum-$cost$ target tree determined by using the proposed cost function.
The target tree in `\textbf{-d}' is the original target tree \cite{feng2018model}.
}
\label{fig:6}
\end{figure*}
The results of the path planning experiments in parking environments with narrow regions are shown in Table \ref{table:2}. 
`\#number-a, b, c, d ($cost$)' represent the target tree algorithms with different $cost$ target tree, integrated with RRT*.
The time for search, $t_{\text{tfs}}$, is the time for determining the minimum-$cost$ target tree by Algorithm \ref{alg:2}.
The time to the first path, $t_{\text{ttfp}}$, means the time for finding the first parking path within the sampling time.
Continuous-curvature target trees (\#number\textbf{-a}, \textbf{b}, \textbf{c}) are shown in Fig. \ref{fig:6}.
Each target tree had a different $cost$ with respect to the length of the straight path.
In the case of \#number\textbf{-a}, the minimum-$cost$ target tree was obtained by Algorithm 2.
For comparison with the original target tree algorithm \cite{feng2018model}, in \#number\textbf{-d}, the original target tree was used with RRT* to plan the parking paths.

\begin{table}[h]
\caption{Experimental results for path planning.
The path length, time for search  ($t_{\text{tfs}}$), time to the first solution path ($t_{\text{ttfp}}$), total planning time ($t_{\text{total}}$), and planning success rate are listed.
}
\centering
\renewcommand{\arraystretch}{1.15} 
\resizebox{0.49\textwidth}{!}
 {\begin{tabular}{| @{ }p{0.35cm}@{} | p{2.7cm} | c | c | c | c | c |} 
 \hline\hline
 &\begin{tabular}{@{}c@{}}Path planning algorithm\end{tabular}
          & \begin{tabular}{@{}c@{}}Path \\ length {[m]}\end{tabular} 
          & \begin{tabular}{@{}c@{}}$t_{\text{tfs}}$ \\ {[ms]}\end{tabular} 
        & \begin{tabular}{@{}c@{}}$t_{\text{ttfp}}$ \\ {[ms]}\end{tabular} 
        & \begin{tabular}{@{}c@{}}$t_{\text{total}}$ \\ {[ms]}\end{tabular} 
          & \begin{tabular}{@{}c@{}}Success \\ rate {[\%]}\end{tabular} \\
 \hline\hline
  \multirow{6}{*}{\rotatebox[origin=c]{90}{perpendicular (\#1)}}
  & \cellcolor{green}\textbf{\#1-a} (\textit{\textbf{0.572}}, min-$cost$) & \textbf{20.6 $\pm$ 2.9} & \textbf{82} $\pm$ \textbf{2} & \textbf{45}
  $\pm$ \textbf{30} & \textbf{127} & \textbf{100}\\
  \cline{2-7}
  &\#1-b (\textit{0.770}) & 25.8 $\pm$ 3.6 & - & 952 $\pm$ 771 & 952 & 75 \\
  \cline{2-7}
  &\#1-c (\textit{0.878})& 28.2 $\pm$ 3.9 & - & 1123 $\pm$ 883 & 1123 & 47 \\
  \cline{2-7}
  &\#1-d (\textit{0.743}) \cite{feng2018model} & 25.6 $\pm$ 3.9 & - & 767 $\pm$ 702 &  767 & 76 \\
  \cline{2-7}
  
  &\begin{tabular}{@{}l@{}}Informed-RRT* \cite{kim2021comparative} \end{tabular} & 27.4 $\pm$ 5.0 & - & 715 $\pm$ 788 &  715 & 84 \\
  \cline{2-7}
  &\begin{tabular}{@{}l@{}}Bidirectional-RRT* \cite{jhang2020autonomous} \end{tabular} & 25.9 $\pm$ 4.7 & - & 737 $\pm$ 698 &  737 & 89 \\
  
 \hline 
 \hline 
  \multirow{6}{*}{\rotatebox[origin=c]{90}{ parallel (\#2)}}
  &\textbf{\#2-a} (\textit{\textbf{0.891}}, min-$cost$) \cellcolor{green}  & \textbf{16.8} $\pm$ \textbf{2.3} & \textbf{12} $\pm$ \textbf{1} & \textbf{21} $\pm$ \textbf{9} & \textbf{33} & \textbf{100}\\
  \cline{2-7}
  &\#2-b (\textit{0.966}) & 19.2 $\pm$ 2.8 & - & 97 $\pm$ 195 & 97 & 100 \\
  \cline{2-7}
  &\#2-c (\textit{0.983})& 21.3 $\pm$ 4.7 & - & 249 $\pm$ 499 & 249 & 96\\
  \cline{2-7}
  &\#2-d (\textit{0.952}) \cite{feng2018model}  & 18.9 $\pm$ 3.1 & - & 51 $\pm$ 89 & 51 & 100\\
  \cline{2-7}
  
  &\begin{tabular}{@{}l@{}} Informed-RRT* \cite{kim2021comparative} \end{tabular} & 21.3 $\pm$ 2.9 & - & 350 $\pm$ 559 &  350 & 89 \\
  \cline{2-7}
  &\begin{tabular}{@{}l@{}} Bidirectional-RRT* \cite{jhang2020autonomous} \end{tabular} & 17.6 $\pm$ 2.8 & - & 161 $\pm$ 168 & 161 & 100 \\
  
 \hline\hline
\end{tabular}}
\label{table:2}
\end{table}

The results show that the target tree which reduces the planning time could be determined by the proposed cost function.
As shown in Table \ref{table:2}, path planning with the minimum-$cost$ target tree (\textbf{-a, proposed}) obtained parking paths with a shorter planning time than did the target trees with higher $costs$ (\textbf{-b, -c, -d}).
In situation \#1-a, in Fig \ref{fig:6}, the parking path was obtained by the RRT* tree reaching the candidate goal of the target tree without searching the narrow region near the parallel-parked vehicle.
Moreover, even though additional planning time was required for determining this minimum-$cost$ target tree by Algorithm \ref{alg:2}, the total planning time, $t_{\text{total}}$, was reduced.

About the target trees with higher $cost$s, the path planning time was increased, and path planning could fail.
As for \#1-b and \#1-c, in Fig. \ref{fig:6}, the RRT tree should be extended toward the target tree without collisions in the narrow region.
This could cause path planning failure within the sampling time, and the path length was increased despite that the parking path was obtained.
Consequently, further sampling time may be required.
In \#1-d, in Fig. \ref{fig:6}, even if the parking path was obtained successfully within the sampling time, the planned path is curvature-discontinuous, so the vehicle could not easily track and park accurately (see Fig. \ref{fig:5}).

Compared with other sampling-based algorithms for parking, the proposed algorithm planned a path within a shorter planning time.
In particular, it reduced the path length and its deviation within the given sampling time by building backward parking paths (target tree) in advance, unlike the bidirectional-RRT* planning algorithm \cite{jhang2020autonomous}, which kept searching for a backward path using the tree built from the goal.

\subsection{Completeness and Optimality Analysis for continuous-curvature target tree algorithm}

This section presents an analysis of whether the probabilistic completeness \cite{lavalle2001randomized} of RRT or RRT* is maintained after the integration of the proposed target tree algorithm.
Additional experiments show that the proposed algorithm can find a path close to the optimal path planned by RRT*, within a shorter sampling time than can the original target tree algorithm and other sampling-based planning algorithms for parking.

In the path planning problem, $\mathcal{P}(q_{init}, q_{goal}, Q_{free})$, the probabilistic completeness of the sampling-based planning algorithm means that if a collision-free path exists, the probability of finding the path will converge to one when the number of random samples approaches to infinity.
The proposed target tree algorithm does not change the completeness of RRT*.
The target tree is a set of poses constituting collision-free paths, and any pose can reach the original goal ($q_{goal}$); i.e., the target tree algorithm is complete with respect to these poses.
In this regard, the target tree, $T_{target}$, can be an extended goal region ($Q_{target}$) that includes not only the original goal ($q_{goal}\in Q_{target}$) but also candidate goals that can reach this original goal.
Accordingly, the proposed algorithm can be deemed to solve the path planning problem to reach this extended goal region with the RRT* path planning algorithm, $\mathcal{P}(q_{init}, Q_{target}, Q_{free})$.
Integration of the target tree algorithm and RRT*, which are both complete, maintains the probabilistic completeness.

\begin{figure}[t!]
\centering
\subfigure[]{\includegraphics[width=0.4925\linewidth]{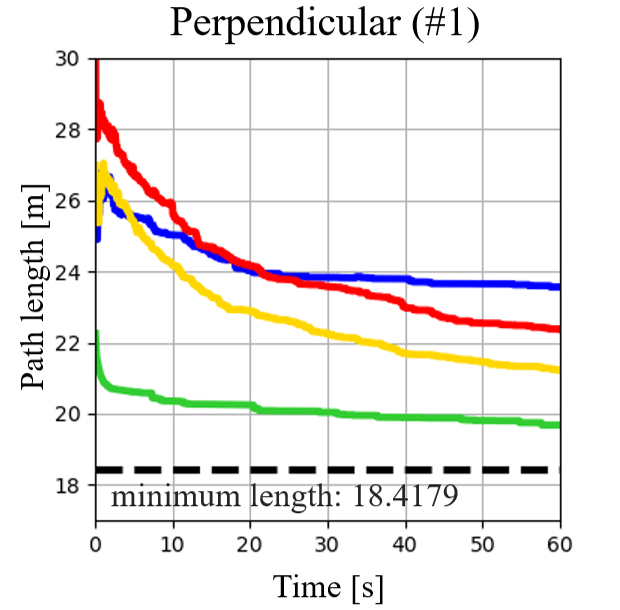}}
\subfigure[]{\includegraphics[width=0.493\linewidth]{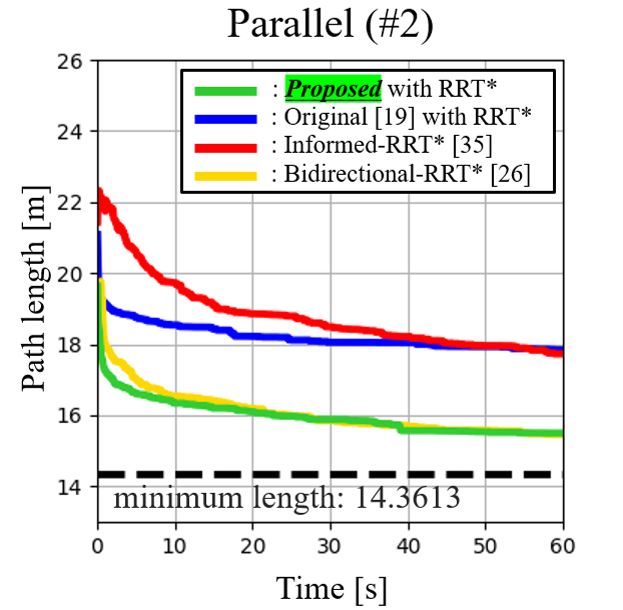}}
\caption{
Comparison results of the mean path length with an increase in the sampling time, under perpendicular and parallel parking (Fig. \ref{fig:6})
}
\label{fig:7}
\end{figure}

Additional experiments were executed to discuss the optimality of the proposed algorithm, as shown in Fig. \ref{fig:7}.
The proposed target tree algorithm was compared with the original target tree algorithm \cite{feng2018model} with RRT* and minimum-length path selection, and other sampling-based planning algorithms for parking \cite{kim2021comparative, jhang2020autonomous}.
Each algorithm was run 100 times for 60 s.
The dashed horizontal line in Fig. \ref{fig:7} is regarded as the minimum length (i.e., optimal cost) in the original RRT* after 7200 s.
The results show that the proposed target tree algorithm can obtain a parking path close to the optimal-length path (i.e., near-optimal).
There are two reasons for these results.
First, integrating RRT* and the proposed minimum-length path selection step (Section \ref{subsec:minimum_length_path_selection}) allows to find a shorter parking path (near-optimal path) as the sampling time increases (see Fig. \ref{fig:7}).
Second, the proposed minimum-$cost$ target tree (Section \ref{subsec:cost_function}) contains backward parking paths similar to the backward path of the optimal path.
For example, as shown in situation \#1-a of Figs. \ref{fig:6} and \ref{fig:7}(a), a shorter parking path was found when planning a path with the proposed target tree considering obstacles.

\section{Conclusion}\label{sec:conclusion}
This paper introduces the continuous-curvature target tree algorithm for complex parking, which addresses the limitations of the original target tree algorithm.
The proposed algorithm uses a continuous-curvature target tree that additionally considers the vehicle's steering velocity.
The algorithm then searches the target tree, thereby possibly reducing the planning time further in complex parking environments.
Integrated with RRT* and minimum-length path selection, the proposed algorithm finds a shorter parking path within a given sampling time.
Experiment results show the practical advantages of the proposed algorithm in real parking environments.
The autonomous vehicle accurately parked, with the cross-track error and lateral/orientation parking alignment error reduced by more than half compared with those of the original target tree algorithm.
In addition, the continuous-curvature path was obtained within relatively short planning times of less than 127 ms for perpendicular parking and 33 ms for parallel parking, and the success rate was 100\%.
In particular, even in complex parking environments, the proposed algorithm found the near-optimal path more rapidly compared with not only the original target tree algorithm but also the informed-RRT* and bidirectional-RRT* path planning algorithms for parking. 
In future work, a target tree that considers the steering acceleration of the vehicle will be studied.

\section*{Acknowledgment}
This work was supported by the research project, Automated Valet Parking Development (No. 490-20200034), funded By Phantom AI Inc. (U.S.).


\ifCLASSOPTIONcaptionsoff
  \newpage
\fi



\bibliographystyle{IEEEtran}
\bibliography{main}
%



%

\begin{IEEEbiography}[{\includegraphics[width=1in,height=1.25in,clip,keepaspectratio]{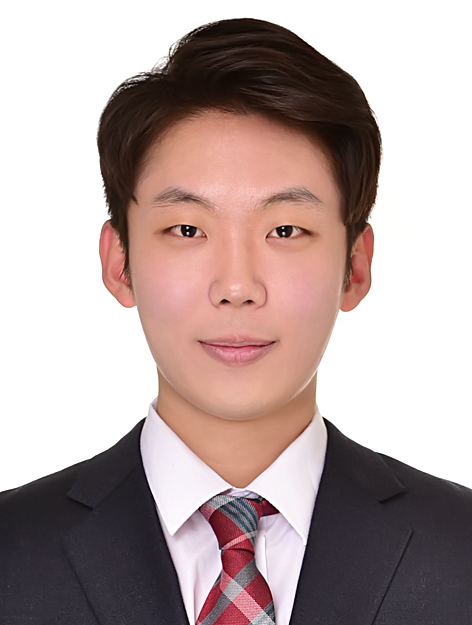}}]{Minsoo Kim}
received the B.S. degree in mechanical engineering from Hanyang University, Seoul, South Korea, in 2018. He is currently pursuing the Ph.D. degree in intelligent systems with the Department of Intelligence and Information, Seoul National University, Seoul. 
His research interests include autonomous vehicle, autonomous valet parking, and motion planning.
\end{IEEEbiography}
\begin{IEEEbiography}[{\includegraphics[width=1in,height=1.25in,clip,keepaspectratio]{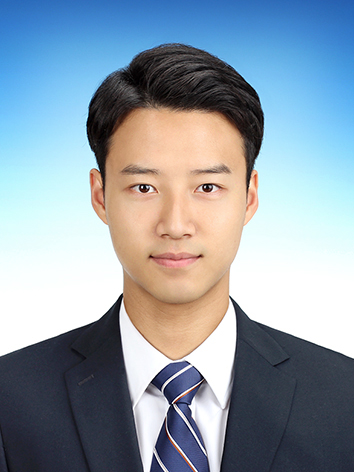}}]{Joonwoo Ahn}
received the B.S. degree in robotics from Kwangwoon University, Seoul, Republic of Korea, in 2016. 
He is currently pursuing the Ph.D. degree in intelligent systems with the Department of Intelligence and Information, Seoul National University, Seoul. His research interests include autonomous valet parking, vision-based navigation, robot-applied deep learning, and path tracking.
\end{IEEEbiography}
\begin{IEEEbiography}[{\includegraphics[width=1in,height=1.25in,clip,keepaspectratio]{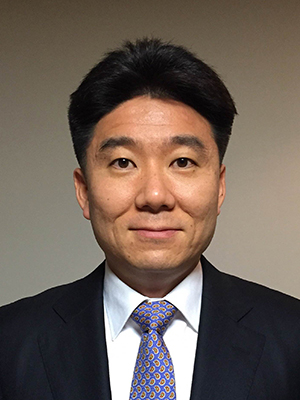}}]{Jaeheung park}
(Member, IEEE) received the B.S. and M.S. degrees in aerospace engineering from Seoul National University, South Korea, in 1995 and 1999, respectively, and the Ph.D. degree in aeronautics and astronautics from Stanford University, CA, USA, in 2006. 
From 2006 to 2009, he was a Postdoctoral Researcher and later a Research Associate with the Stanford Artificial Intelligence Laboratory. 
From 2007 to 2008, he worked part-time with Hansen Medical Inc., USA, a medical robotics company. 
Since 2009, he has been a Professor with the Graduate School of Convergence Science and Technology, Seoul National University. 
His research interests include robot-environment interaction, contact force control, robust haptic teleoperation, multicontact control, whole-body dynamic control, biomechanics, autonomous vehicle, and medical robotics.
\end{IEEEbiography}





\end{document}